\documentclass[12pt]{article}

\usepackage[
  margin=1.5cm,
  includefoot,
  footskip=30pt,
]{geometry}
\usepackage[table]{xcolor}
\usepackage{tikz}
\usetikzlibrary{matrix, arrows}
\usepackage{amsmath,amssymb}
\usepackage{amsthm}
\usepackage{mathtools}
\usepackage{xspace}
\usepackage[noend]{algorithmic}
\usepackage[ruled,vlined]{algorithm2e}
\usepackage{url}
\usepackage{makeidx}
\usepackage{enumerate}
\usepackage{epstopdf}
\usepackage{booktabs}
\usepackage{color}
\usepackage[utf8]{inputenc}
\usepackage{thm-restate}
\usepackage{scalerel,stackengine}
\usepackage[shortlabels]{enumitem}
\usepackage{xr}
\usepackage{fancyvrb}
\usepackage{bold-extra}
\usepackage[width=474.18663pt]{caption}
\usepackage{subcaption}
\usepackage[most]{tcolorbox}
\usepackage{fvextra}
\usepackage{float}
\usepackage{alltt}
\usepackage{soul}
\usepackage{fancyvrb}
\usepackage{multirow}
\usepackage{array}
\usepackage{threeparttable}
\usepackage{authblk}
\usepackage{multicol}
\usepackage{xcolor}
\usepackage{colortbl}
\usepackage{siunitx}
\usepackage{fancyvrb}
\usepackage{pdfpages} 

\usepackage[authoryear,round,sort&compress]{natbib}
\usepackage[final]{hyperref}
\usepackage[bottom]{footmisc}

\usepackage{booktabs,tabularx}
\usepackage{enumitem}   %
\usepackage{parskip}    %
\usepackage{tcolorbox}  %
\tcbset{
  enhanced,
  breakable,          %
  colback=black!3,    %
  colframe=black!60,  %
  boxrule=0.6pt,      %
  arc=2pt,            %
}

\usepackage[activate={true,nocompatibility},final,tracking=true,kerning=true,spacing=true,factor=1100,stretch=10,shrink=10]{microtype}

\usepackage{listings}    %
\usepackage{enumitem}    %

\newcommand{\name}{Sequential Diagnosis Benchmark}
\newcommand{\nameShort}{SDBench}
\newcommand{\mai}{MAI-DxO}
\newcommand{\mailong}{MAI-Dx Orchestrator}

\newcommand{\equalcontrib}{\textsuperscript{*}}

\usepackage{listings}
\usepackage{tikz}
\usetikzlibrary{shapes,calc,positioning}

\global

\tcbset{
  aibox/.style={
    width=474.18663pt,
    top=10pt,
    colback=white,
    colframe=black,
    colbacktitle=black,
    enhanced,
    center,
    attach boxed title to top left={yshift=-0.1in,xshift=0.15in},
    boxed title style={boxrule=0pt,colframe=white,},
  }
}
\newtcolorbox{AIbox}[2][]{aibox,breakable=false,title=#2,#1}

\definecolor{Gray}{gray}{0.95}
\definecolor{aigold}{RGB}{244,210, 1} 
\definecolor{aigreen}{RGB}{210,244,211} 

\sethlcolor{aigreen}

\definecolor{aired}{RGB}{255,180,181}

\newtcbox{\mybox}[1][green]{on line,
arc=0pt,outer arc=0pt,colback=#1!10!white,colframe=#1!50!black,
boxsep=0pt,left=0pt,right=0pt,top=0pt,bottom=0pt,
boxrule=0pt,bottomrule=0pt,toprule=0pt}

\title{Sequential Diagnosis with Language Models}

\author[]{%
  Harsha Nori\equalcontrib, %
  Mayank Daswani\equalcontrib, %
  Christopher Kelly\equalcontrib, %
   \authorcr %
  Scott Lundberg\equalcontrib, 
  Marco Tulio Ribeiro\equalcontrib, %
  Marc Wilson\equalcontrib,
   \authorcr %
  Xiaoxuan Liu, 
  Viknesh Sounderajah,
  Jonathan M Carlson,
  Matthew P Lungren, \authorcr%
  Bay Gross, %
  Peter Hames, %
  Mustafa Suleyman,
  Dominic King, %
  Eric Horvitz}

\affil{Microsoft AI}
\begin{document}

\maketitle
    
\begingroup
  \renewcommand\thefootnote{\fnsymbol{footnote}} %
  \footnotetext[1]{Equal contribution}
  \footnotetext[0]{Correspondence: \texttt{hanori@microsoft.com}, \texttt{horvitz@microsoft.com}}
\endgroup

\begin{abstract}
\noindent
Artificial intelligence holds great promise for expanding access to expert medical knowledge and reasoning.
However, most evaluations of language models rely on static vignettes and multiple-choice questions that fail to reflect the complexity and nuance of evidence-based medicine in real-world settings.
In clinical practice, physicians iteratively formulate and revise diagnostic hypotheses, adapting each subsequent question and test to what they’ve just learned, and weigh the evolving evidence before committing to a final diagnosis.
To emulate this iterative diagnostic process, we introduce the \mbox{\emph{\name{}}}, which transforms 304 diagnostically challenging New England Journal of Medicine clinicopathological conference (NEJM-CPC) cases into stepwise diagnostic encounters.
A physician or AI begins with a short case abstract and must iteratively request additional details from a gatekeeper model that reveals findings only when explicitly queried.
Performance is assessed not just by diagnostic accuracy but also by the cost of physician visits and tests performed.
To complement the benchmark, we present the \emph{MAI Diagnostic Orchestrator (\mai{})}, a model-agnostic orchestrator that simulates a panel of physicians, proposes likely differential diagnoses and strategically selects high-value, cost-effective tests.
When paired with OpenAI’s o3 model, \mai{} achieves 80\% diagnostic accuracy—four times higher than the 20\% average of generalist physicians. \mai{} also reduces diagnostic costs by 20\% compared to physicians, and 70\% compared to off-the-shelf o3.
When configured for maximum accuracy, \mai{} achieves 85.5\% accuracy.
These performance gains with \mai{} generalize across models from the OpenAI, Gemini, Claude, Grok, DeepSeek, and Llama families.
We highlight how AI systems, when guided to think iteratively and act judiciously, can advance both diagnostic precision and cost-effectiveness in clinical care.
\end{abstract}

\newpage

\section{Introduction}

\emph{Sequential diagnosis} is a cornerstone of clinical reasoning, wherein physicians refine their diagnostic hypotheses step-by-step through iterative questioning and testing.
Figure \ref{fig:1-gameplay} illustrates how a diagnostician might approach a case given limited initial information, posing broad then increasingly specific questions to narrow down the differential to a likely malignancy, followed by imaging, biopsy, and specialist studies to arrive at a final diagnosis. Solving such cases demands a complementary set of skills:  identifying the most informative next questions or tests, balancing marginal diagnostic yield against cost and patient burden, and recognizing when the evidence is sufficient to make a confident diagnosis.

Language models (LMs) have demonstrated impressive diagnostic capabilities, with recent studies showing top-tier performance on medical licensing exams and highly structured diagnostic vignettes \citep{nori2023capabilities, nori2023can, mcduff2025towards, cabral2024clinical, nori2024medprompt, goh2024large}. However, these evaluations occur under artificial conditions that differ markedly from real-world clinical practice. Most diagnostic assessments present models with neatly packaged vignettes that bundle the chief complaint, history of present illness, key physical exam findings, and test results, and then ask the model to select a diagnosis from a predefined answer set. By reducing the sequential diagnosis cycle to a one-turn multiple-choice quiz, static benchmarks risk overstating model competence and obscure potential weaknesses including premature diagnostic closure, indiscriminate test ordering, and anchoring on early hypotheses.

We introduce the \textbf{\name{}} (\textbf{\nameShort{}}), an interactive framework for evaluating diagnostic agents (human or AI) through realistic sequential clinical encounters. \nameShort{} recasts 304 New England Journal of Medicine (NEJM) clinicopathological conference (CPC) cases into stepwise diagnostic encounters in which a \emph{diagnostic agent} decides which questions to ask, which tests to order, and when to commit to a final diagnosis. Information is revealed by an information \emph{Gatekeeper}, a language model that serves as an oracle for the patient case. The Gatekeeper discloses specific clinical findings only when explicitly queried, and can synthesize additional case-consistent information for tests not described in the original CPC narrative. Once a final diagnosis is submitted, we evaluate its correctness against the ground truth diagnosis, and compute the cumulative estimated real world cost of all requested diagnostic tests. By measuring both diagnostic accuracy \emph{and} cost, \nameShort{} aligns with the goals of the Triple Aim \citep{tripleaim}, which seeks high quality care delivered at sustainable cost. 
A cohort of U.S. and U.K. physicians with a median of 12 years of experience achieved 20\% accuracy at an average cost of \$2,963 per case on \nameShort{}, underscoring the inherent difficulty of the benchmark. Off-the-shelf commercial models showed varied tradeoffs: GPT-4o achieved 49.3\% accuracy at a lower cost (\$2,745 per case), while o3 reached 78.6\% accuracy at substantially higher cost (\$7,850 per case).

We further introduce \textbf{MAI Diagnostic Orchestrator (\mai{})}, an orchestrated system co-designed with physicians that consistently outperforms both human physicians and commercial language models along the cost-accuracy Pareto frontier. Compared to off-the-shelf LMs, \mai{} improves diagnostic accuracy while cutting estimated medical costs by more than half, demonstrating the power of careful orchestration even atop state-of-the-art models. For instance, while the off-the-shelf o3 model achieved 78.6\% accuracy at a cost of \$7,850, \mai{} achieved 79.9\% at just \$2,397, or 85.5\% at \$7,184 (Section \ref{sec:results}).
These gains stem from a set of physician-inspired strategies: simulating a virtual panel of physicians with distinct roles, estimating marginal costs between diagnostic rounds, and employing model ensembling methods across model responses. Crucially, these techniques are general-purpose: \mai{} boosted the accuracy of off-the-shelf models from a variety of providers by an average of $11$ percentage points.

In summary, our contributions bring AI-driven diagnosis closer to clinical utility on two key fronts.
First, \nameShort{} transcends static benchmarks by aligning with the dynamic, uncertain nature of real-world diagnostic reasoning. Prior work using NEJM CPCs for assessing diagnostic reasoning \citep{mcduff2025towards, brodeur2024superhuman} presented the full case upfront and asked for the top-k diagnoses---implicitly assuming perfect information. In contrast, \nameShort{} challenges diagnostic agents to decide which questions or tests to request, in what order, and when to commit to a final diagnosis, all under cost constraints. 
This allows us to assess not only diagnostic accuracy, but also an agent’s ability to seek the most informative evidence in a cost-conscious manner, and to recognize when diagnostic certainty is warranted.
Second, \mai{} shows what is \emph{already achievable} with thoughtful orchestration of today's best off-the-shelf models, surpassing experienced physicians by 4x on accuracy while also \emph{reducing cost}. 
Together, \nameShort{} and \mai{} establish an empirically grounded foundation for advancing AI-assisted diagnosis under realistic constraints.

\section{Sequential Diagnosis Benchmark}
\label{sec:overview}

\begin{figure}[H]
    \centering
    \includegraphics[width=0.95\linewidth]{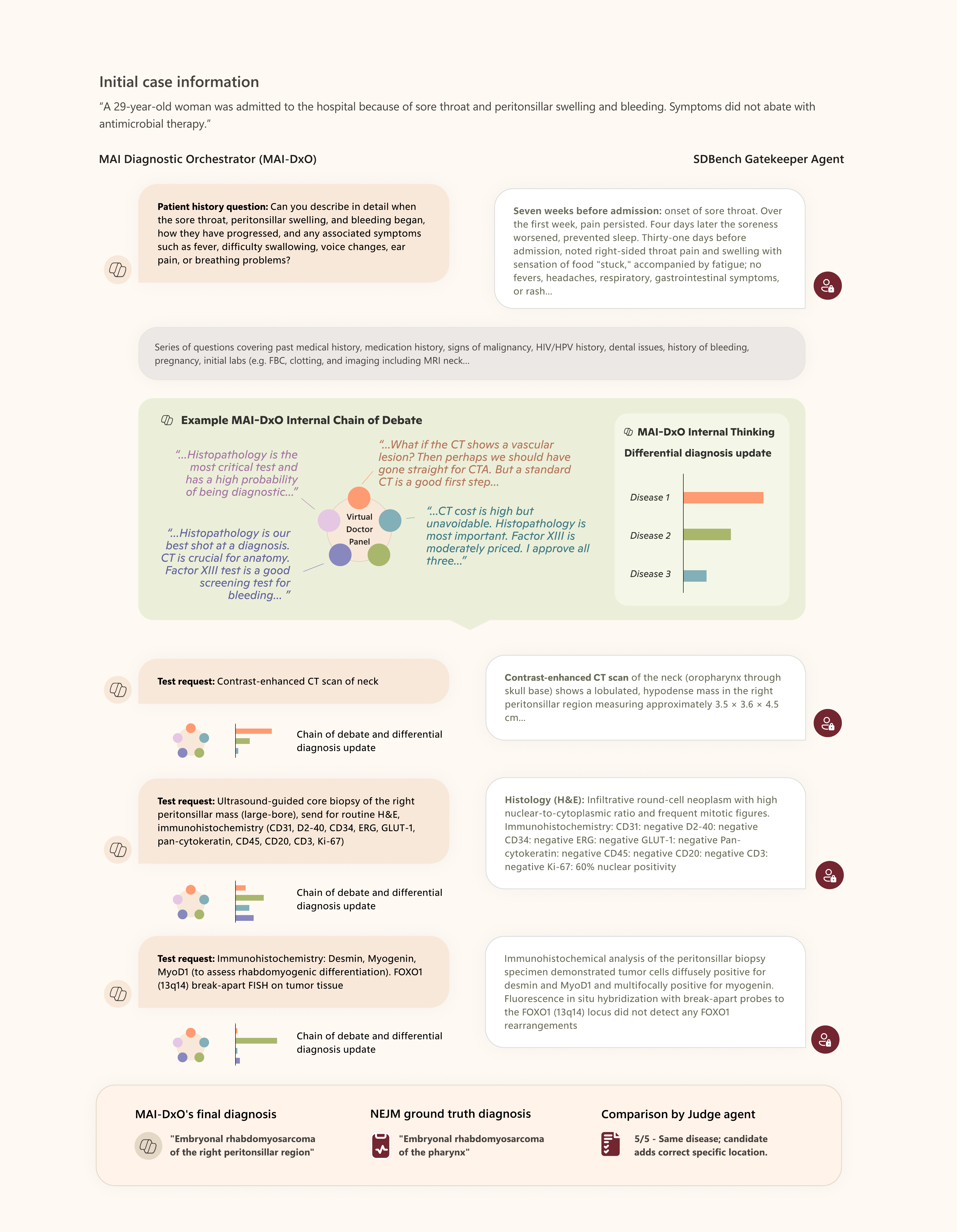}
    \caption{Example of an AI agent solving a sequential-diagnosis reasoning problem.}
    \label{fig:1-gameplay}
\end{figure}

In order to build the \name{} (\nameShort{}), we took cases from the New England Journal of Medicine’s (NEJM) Case Challenge series. The data set spans a diverse array of clinical presentations, with final diagnoses ranging from common conditions (e.g., ``Covid-19 pneumonia'') to rare disorders (e.g., ``Neonatal hypoglycaemia due to a biologically active teratoma").
We collected 304 consecutive cases published between 2017 and 2025, converting each into an interactive simulation of sequential diagnostic reasoning. Each encounter begins with a brief summary of the patient and their chief complaint, for example: ``A 29-year-old woman was admitted to the hospital because of sore throat and peritonsillar swelling and bleeding. Symptoms did not abate with antimicrobial therapy'' (Figure \ref{fig:1-gameplay}). From that starting point, a diagnostic agent (or human physician) may take one of the following actions: 
\begin{enumerate}
    \item \textbf{Ask questions:} free-text questions for history or examination details (``Has she traveled recently?''). Multiple questions are allowed.
    
    \item \textbf{Request diagnostic tests:} explicit orders for labs, imaging, or procedures (``Order a CT chest with contrast'').
    
    \item \textbf{Diagnose:} a one-time commitment to a final diagnosis (``The diagnosis is histoplasmosis.'').
\end{enumerate}

The Gatekeeper agent (described in detail below) interprets each request, consults the full case file, and responds in plain language, either providing the requested information or issuing a refusal if the query is too vague or non-specific. When the Diagnostic agent chooses the `diagnose' action, the Judge evaluates the proposed diagnosis for correctness, and a Cost Estimator calculates the total expense of all tests ordered. The Diagnostic Agent is evaluated along two axes: diagnostic accuracy and cumulative testing cost.

\paragraph{Gatekeeper.}
We implemented the Gatekeeper using a language model (o4-mini) with access to the full NEJM CPC case file, including the final diagnosis. Guided by physician-devised rules, the Gatekeeper discloses only information that a real-world clinician could legitimately obtain from a given query or test, such as specific test results, succinct patient-history, or physical exam findings. It explicitly refuses to provide diagnostic impressions, interpret test results, or offer hints that would be unavailable in a genuine clinical encounter. Imaging is withheld until explicitly ordered; pathognomonic findings are disclosed only when the exact confirmatory test is requested; and vague or overly broad requests trigger polite refusals. Direct questions about the patient's history or examination return responses in clinical language, closely mirroring the information extraction task faced by physicians when reviewing a medical record. Figure \ref{fig:1-gameplay} illustrates sample requests and responses. Through this approach, the Gatekeeper removes spoilers and hindsight bias commonly embedded in educational case write-ups.

In early pilot studies with physicians and LMs, we observed that a significant fraction of information requests targeted patient details or test results not present in the original published cases. Our initial strategy of responding ``Not Available" had unintended side effects: it implicitly signaled which queries were off-path and discouraged valid alternative clinical reasoning pathways. To address this, we changed the Gatekeeper to return realistic \emph{synthetic findings} for queries not covered in the original text. These findings are numerically or descriptively consistent with the rest of the case, with no indication that they are synthetic.
By returning what would \emph{likely} have been found had the test been performed, the Gatekeeper preserves clinical realism while avoiding implicit clues from missing data.

We further validated the Gatekeeper’s behavior by asking a panel of physicians to review $508$ Gatekeeper responses, comprising both real and synthetic outputs. Reviewers were instructed to search for and categorize any inappropriate responses, including clues that could ``leak" diagnostic information, findings from tests not ordered, clinical interpretations beyond objective test results, and pathognomonic results offered prematurely. Reviewers flagged only eight responses as potentially problematic, and none were judged to have leaked the diagnosis after group adjudication.

\begin{figure}[t]
    \centering
    \includegraphics[width=0.9\linewidth]{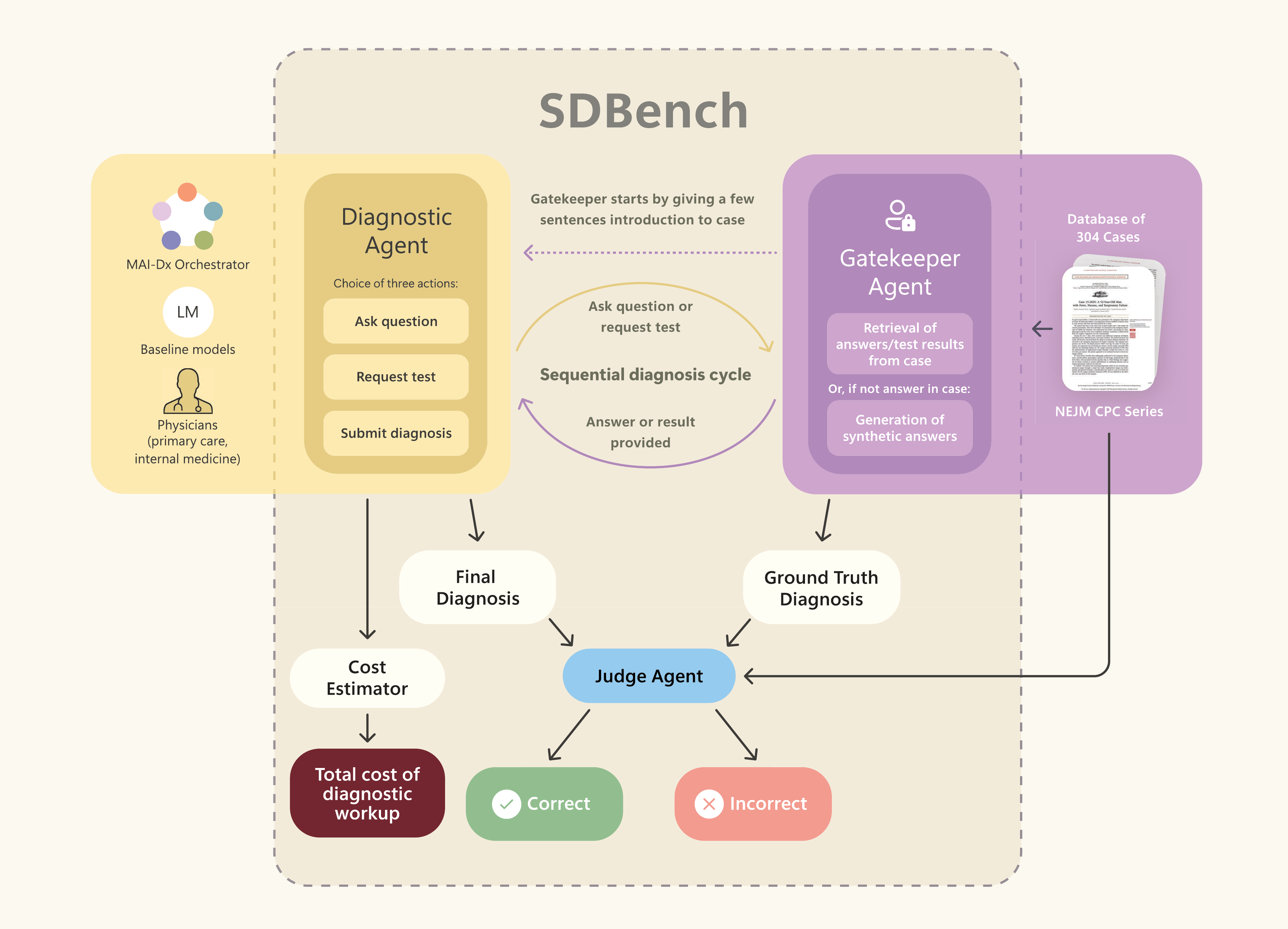}
    \caption{Multiagent orchestration in the \nameShort{} benchmark. A corpus of NEJM CPC cases is transformed into sequential diagnosis challenges through coordination among three agents: the Gatekeeper, Diagnostic and Judge agents. At run-time, the Gatekeeper mediates requests for information from the Diagnostic agent, deciding if and how to respond to the Diagnostic agent's questions about patient history, examination findings, and test results. The Judge evaluates whether the Diagnostic Agent's final diagnosis matches the ground truth reported in the original CPC article.}
    \label{fig:architecture}
\end{figure}

\paragraph{Judging diagnoses against ground truth.}
Two physicians may reasonably describe the same condition using different terminology, e.g. ``bacterial endocarditis'' versus ``infective endocarditis due to Staphylococcus aureus'', yet arrive at identical treatment decisions. To account for such variability, we introduced a Judge agent to evaluate diagnoses based on clinical substance rather than surface-form descriptions. The Judge was implemented using the o3 model prompted with a detailed, physician-authored rubric (Table \ref{tab:judgerubric}) designed to reflect clinical consensus, similar in spirit to ~\citet{arora2025healthbench}. The rubric evaluates key dimensions of diagnostic quality, including the core disease entity, etiology, anatomic site, specificity, and overall completeness, with a particular emphasis on whether the candidate diagnosis would meaningfully alter clinical management. To ensure contextual understanding, the Judge had full access to each case file during adjudication. We set a cut-off of $\geq 4$ on a five-point Likert scale to count as a “correct” diagnosis, based on the clinical rationale that clinical management would remain largely unchanged above this threshold.

To validate the Judge, in-house physicians independently graded all 56 test-set diagnoses produced by our most accurate Diagnostic Agent (see next section), as well as 56 randomly selected human-generated diagnoses (one for each case). After binarizing both the physician and Judge scores, we found that inter-rater agreement was strong - Cohen’s $\kappa$ = 0.70 for the MAI-Dx set and $\kappa$ = 0.87 for the human set. In four out of the five (total) cases of disagreements, physicians judged the automated Judge to be overly strict, marking correct diagnoses as incorrect.

\begin{table}[t]
\centering
\begin{tabularx}{\linewidth}{@{}c l X@{}}
\toprule
\textbf{Score} & \textbf{Label} & \textbf{Definition / Anchor} \\ \midrule
\textbf{5} & Perfect / Clinically superior
  & Clinically identical to the reference \textbf{or} a strictly more specific version.  
    Any added detail must be \emph{directly related} (complication, organ infiltration, sequelae).  
    No unrelated or incorrect additions. \\[3pt]

\textbf{4} & Mostly correct (minor incompleteness)
  & Core disease correctly identified but a secondary qualifier or component is missing or slightly mis-specified.  
    Overall management would remain largely unchanged. \\[3pt]

\textbf{3} & Partially correct (major error)
  & Correct general disease category, but a major error in etiology, site, or critical specificity \emph{or} inclusion of an unrelated diagnosis alongside a correct one.  
    Would alter work-up or prognosis. \\[3pt]

\textbf{2} & Largely incorrect
  & Shares superficial features only (e.g.\ manifestation without etiology, different disease in same category).  
    Fundamentally misdirects clinical work-up or partially contradicts case details. \\[3pt]

\textbf{1} & Completely incorrect
  & No meaningful overlap; wrong organ/system; nonsensical or contradictory additions.  
    Following this diagnosis would likely lead to harmful care. \\ \bottomrule
\end{tabularx}
\caption{Five-point Likert rubric used by the \textit{Judge} agent.  
Each score is assigned after comparing the candidate diagnosis with the reference diagnosis across
(1) core disease entity, (2) etiology/cause, (3) anatomic site, (4) specificity/qualifiers, and (5) completeness.  
Accepted medical synonyms (e.g.\ “Hodgkin lymphoma” vs.\ “Hodgkin’s disease”) are considered equivalent.}
\label{tab:judgerubric}
\end{table}

\paragraph{Estimating costs.}
Using monetary cost as a secondary evaluation metric helps to deter unrealistic diagnostic behaviors, such as the indiscriminate ordering of imaging or biopsies. In everyday clinical practice, the potential diagnostic yield of each investigation must be weighed against practical constraints like monetary costs, invasiveness, time to result, and insurance limitations. Since monetary cost often reflects these real-world constraints, it serves as a helpful proxy for these multifaceted factors.

We treat sequential patient-facing questions and physical examination findings as part of a standard physician visit, assigning a fixed cost of \$300 per visit. Diagnostic test costs were determined using a language model-based lookup system designed to translate diagnostic test requests, provided in free-text format, into standardized Current Procedural Terminology (CPT) codes. For more complex diagnostic investigations, the system was able to assign multiple CPT codes. These CPT codes were then matched to corresponding cost data derived from a 2023 pricing table published by a large U.S. health system, sourced under the CMS HHS price transparency rule (45 CFR §180). Our system was able to match requested tests to relevant CPT codes over 98\% of the time; in the remaining edge cases, we used a language model to estimate a price. Although the resulting cost estimates are not intended to be exact representations of actual clinical expenses, they offer a standardized and consistent approach to comparatively assess costs across different diagnostic agents and physicians.

\section{Experimental Setup}

We evaluated both physicians and diagnostic agents on the 304 NEJM Case Challenge cases in \nameShort{}, spanning publications from
2017 to 2025. The most recent $56$ cases (from 2024--2025) were held out as a hidden test set to assess generalization performance. These cases remained unseen during development. We selected the most recent cases in part to assess for potential memorization, since many were published after the training cut-off dates of the language models under evaluation. 

As described in Section \ref{sec:overview}, each case begins with a brief
clinical vignette (typically 2--3 sentences, as in Figure \ref{fig:starting-info}) summarizing the patient's chief
complaint. From this starting point, diagnostic agents interact
with the Gatekeeper in a sequence of turns until they reach a diagnosis. At each turn, the agent may: (i) ask about patient history or physical examination findings, (ii) order a diagnostic test, or (iii) commit to a final diagnosis.

\begin{figure}[H]
\begin{AIbox}{Initial Information Provided}

\begin{verbatim}
A 52-year-old man was evaluated in a hospital in Argentina for fever
and hypoxemic respiratory failure. Imaging showed pulmonary opacities, 
and the hematocrit was 56.9%
\end{verbatim}
\end{AIbox}
\caption{Participating physicians and models are provided with a case abstract to begin the sequential diagnosis process. Example drawn from May 28, 2025 NEJM CPC case \citep{doi:10.1056/NEJMcpc2412526}.}
\label{fig:starting-info}
\end{figure}

The evaluation protocol mirrors real clinical constraints: agents must specify tests explicitly (e.g., ``CBC'' rather than ``blood work'') and incur cumulative costs for each ordered test. As noted before, the Gatekeeper synthesized
plausible results when the requested tests were not mentioned in the original case to prevent information leakage while maintaining clinical realism. 
Our primary metric was diagnostic accuracy, defined as the percentage of cases
receiving a judge score $\geq 4$ on the 5-point clinical accuracy rubric, 
corresponding to diagnoses that would lead to appropriate treatment. We used cost as a secondary metric, computed as the cost (in USD) of all diagnostic tests ordered before reaching a diagnosis, plus a fixed cost per physician visit (\$300). Multiple sequential questions constituted one physician visit, which concluded upon a diagnostic test request. 

\subsection{Off-the-shelf models}

We also evaluated a comprehensive suite of state-of-the-art language models spanning multiple model families and sizes, from a variety of model providers. Foundation models tested include: GPT-3.5-turbo, GPT-4o, GPT-4.1, GPT-4.1-mini, GPT-4.1-nano, o3, o4-mini, Claude 4 Sonnet, Claude 4 Opus, Gemini 2.5 Pro, Gemini 2.5 Flash, Grok-3, Grok-3-mini, Llama 4 Maverick, and Deepseek-R1.

For baseline evaluations, we employed a minimal prompt designed to test ``out-of-the-box'' diagnostic capabilities.  The baseline prompt (Figure~\ref{fig:baseline-prompt}) instructed models to use simple XML tags for requesting tests (\texttt{<test>}) and asking questions (\texttt{<question>}), with a final \texttt{<diagnosis>} tag for submitting their answer. This straightforward format provided a basis for fair comparison across model families.

\begin{figure}[H]
\begin{AIbox}{Baseline Performance Prompt}
\begin{verbatim}
You are a diagnostic assistant. Order tests and ask patient questions 
to determine the diagnosis.

To order tests use <test></test> tags:
<test>CBC</test>
<test>Chest X-ray</test>
...more tests...

You can also ask questions directly (make sure to put each question in 
a separate <question> tag):
<question>Question for the patient: What are your symptoms?</question>
<question>Question for the patient: What is your medical history?
</question>...more questions...

You cannot mix <test> and <question> tags in the same turn, just use all 
<test> tags or all <question> tags. 
Make sure to ask for enough questions and tests to reach a diagnosis.

When ready to diagnose, use <diagnosis></diagnosis> tags:
<diagnosis>Your diagnosis here</diagnosis>
\end{verbatim}
\end{AIbox}
\caption{Prompt used for baseline performance estimation.}
\label{fig:baseline-prompt}
\end{figure}

\subsection{MAI Diagnostic Orchestrator}
\begin{figure}[H]
    \centering
    \includegraphics[width=\linewidth]{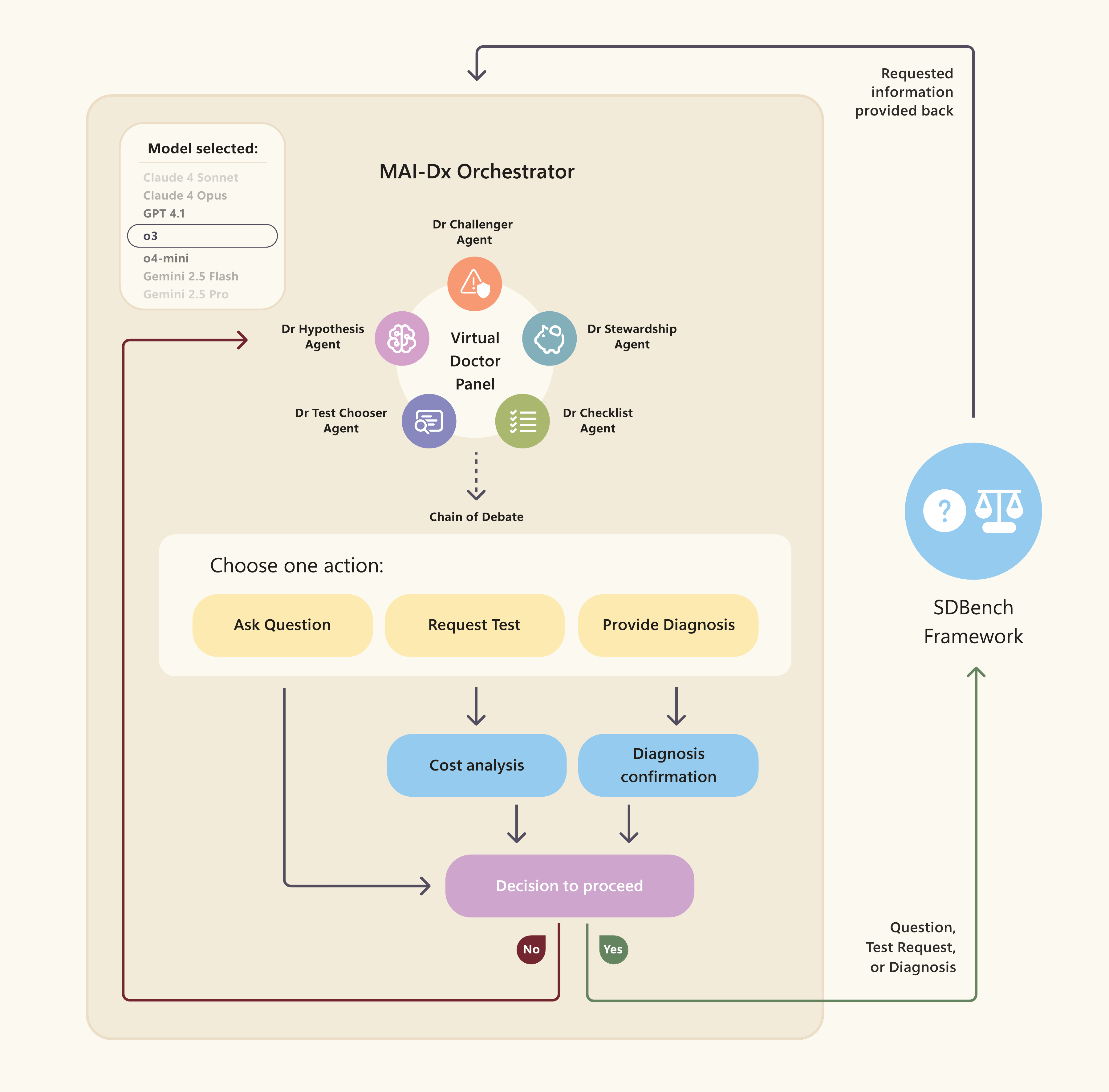}
    \caption{Overview of the \mailong{}}
    \label{fig:MAI-DxO Architecture}
\end{figure}

With input from physicians, we developed a system that emulates a virtual panel of doctors working collaboratively to solve diagnostic cases (\mai{}).
As shown in Figure~\ref{fig:MAI-DxO Architecture}, a single language model role-plays five distinct medical personas, each contributing specialized expertise to the diagnostic process. This orchestrated approach aims to replicate the benefits of team-based clinical reasoning while mitigating individual cognitive biases and minimizing cost/invasiveness.

The virtual panel consists of five specialized roles:
\begin{itemize}
    \item \textbf{Dr. Hypothesis} -- Maintains a probability-ranked differential diagnosis with the top three most likely conditions, updating probabilities in a Bayesian manner after each new finding.
    \item \textbf{Dr. Test-Chooser} -- Selects up to three diagnostic tests per round that maximally discriminate between leading hypotheses
    \item \textbf{Dr. Challenger} -- Acts as devil's advocate by identifying potential anchoring bias, highlighting contradictory evidence, and proposing tests that could falsify the current leading diagnosis.
    \item \textbf{Dr. Stewardship} -- Enforces cost-conscious care by advocating for cheaper alternatives when diagnostically equivalent and vetoing low-yield expensive tests.
    \item \textbf{Dr. Checklist} -- Performs silent quality control to ensure the model generates valid test names and maintains internal consistency across the panel's reasoning.
\end{itemize}

After a \textbf{``Chain of Debate"} -- a structured deliberation between sets of agents with defined roles --  the panel reaches consensus on one of three actions: asking questions, ordering tests, or committing to a diagnosis (if certainty exceeds threshold). Before tests are ordered, an optional \emph{budget tracker} can be invoked to estimate both the cumulative medical costs so far and the cost of each test in the order.

We evaluate five variants of \mai{} to explore different points on the accuracy-cost frontier (from most cost conscious to least):
\begin{itemize}
    \item \textbf{Instant Answer} -- Diagnosis based solely on initial vignette (as in Figure \ref{fig:starting-info}), without any follow-up questions or tests.
    \item \textbf{Question Only} -- The panel can ask questions, but cannot order diagnostic tests. The cost is simply the cost of a single physician visit.
    \item \textbf{Budgeted} -- The panel is augmented with a budgeting system that tracks cumulative costs (a separately orchestrated language model call) towards a max budget and allows the panel to cancel tests after seeing their estimated cost.
    \item \textbf{Unconstrained Budget} -- Full panel with no explicit cost tracking or budget limitations.
    \item \textbf{Ensemble} -- Simulates multiple doctor panels working in parallel, with an additional panel to provide a final diagnosis. This is implemented as multiple independent No Budget runs with a final aggregation step to select the best diagnosis. Costs are computed as the sum of the costs of all tests ordered by each of the runs, accounting for duplicates.
\end{itemize}

\mai{} was primarily developed and optimized using GPT-4.1, but is designed to be model-agnostic. %
All \mai{} variants used the same underlying orchestration structure, with capabilities selectively enabled or disabled for variants.

\subsection{Physicians}

\begin{figure}
    \centering
    \includegraphics[width=\linewidth]{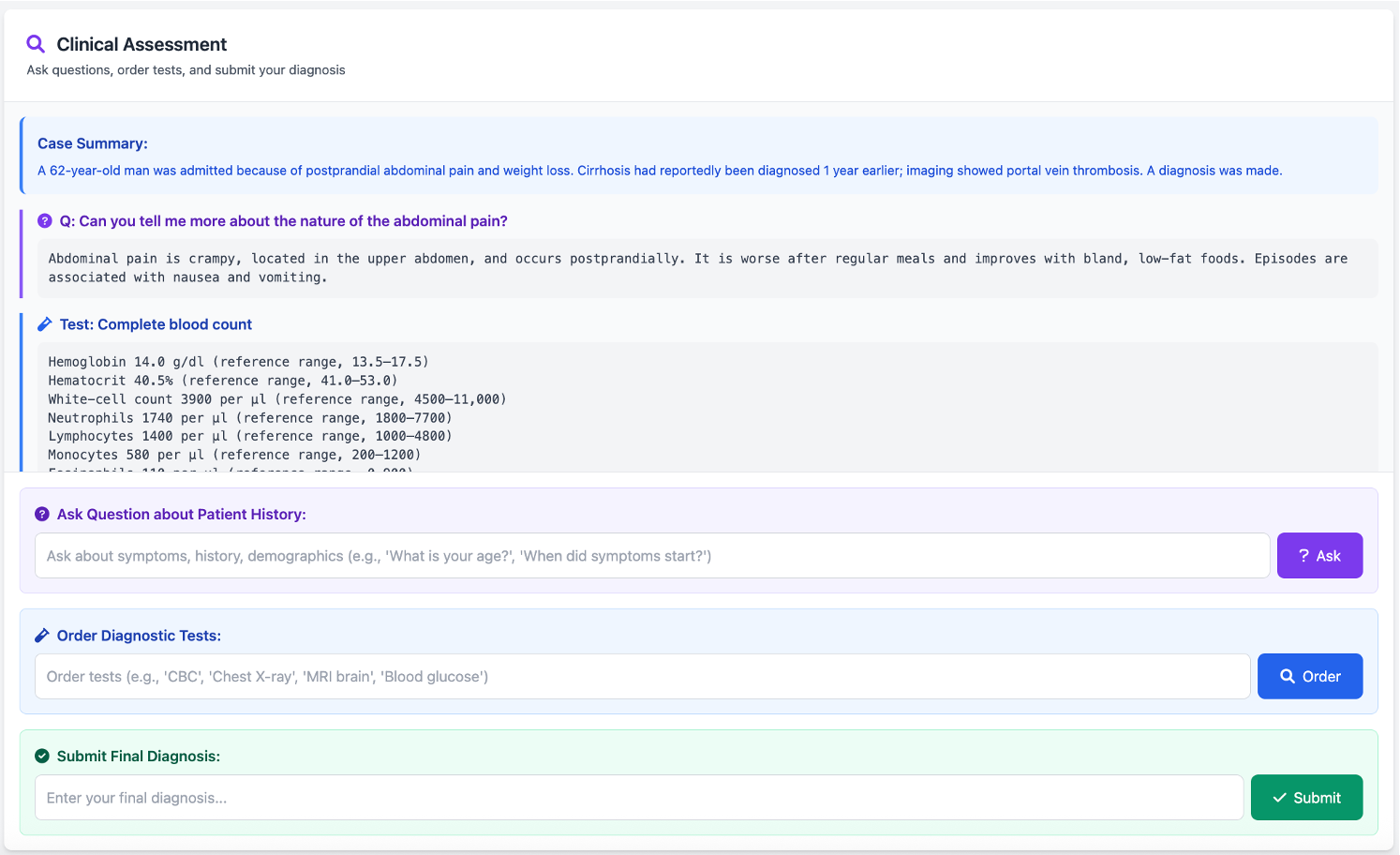}
    \caption{Interface developed for physicians to attempt cases from \nameShort{}.}
    \label{fig:game-ui}
\end{figure}

To assess the relative performance of AI agents and practicing physicians, 
we developed a synchronous text-chat user interface that allows a human user to assume the role of the diagnostic agent and converse with the Gatekeeper model in order to ask questions, request diagnostic tests and, ultimately, provide a differential diagnosis (Figure \ref{fig:game-ui}). Thus, human physicians participated in \nameShort{} the same way as an AI diagnostic agent.

To establish human performance, we recruited $21$ physicians practicing in the US or UK to act as diagnostic agents. Participants had a median of 12 years [IQR 6-24 years] of experience: 17 were primary care physicians and four were in-hospital generalists.
Each physician received the same initial vignette as the AI agents, and interacted with an identical Gatekeeper interface. No limits were imposed on session duration or the number of tests ordered.
Cases were drawn from the hidden test set and case order was randomized for each participant to mitigate ordering effects. Physicians were blinded to the correctness of their diagnosis, and were asked to complete as many as possible during the study period.

Physicians were explicitly instructed not to use external resources, including search engines (e.g., Google, Bing), language models (e.g., ChatGPT, Gemini, Copilot, etc), or other online sources of medical information. Although limiting the use of search engines may not accurately reflect physicians’ real world clinical practice, the original NEJM cases are accessible online, and we sought to prevent participants from readily obtaining correct answers through external searches. Additionally, certain search engines offer AI-generated summaries, potentially providing diagnostic hints. By restricting physicians’ access to language models, we aimed specifically to assess their intrinsic diagnostic capabilities, rather than indirectly evaluating the performance of available generative artificial intelligence tools.

\section{Results}
\label{sec:results}
\begin{figure}[H]
    \centering
    \includegraphics[width=\linewidth]{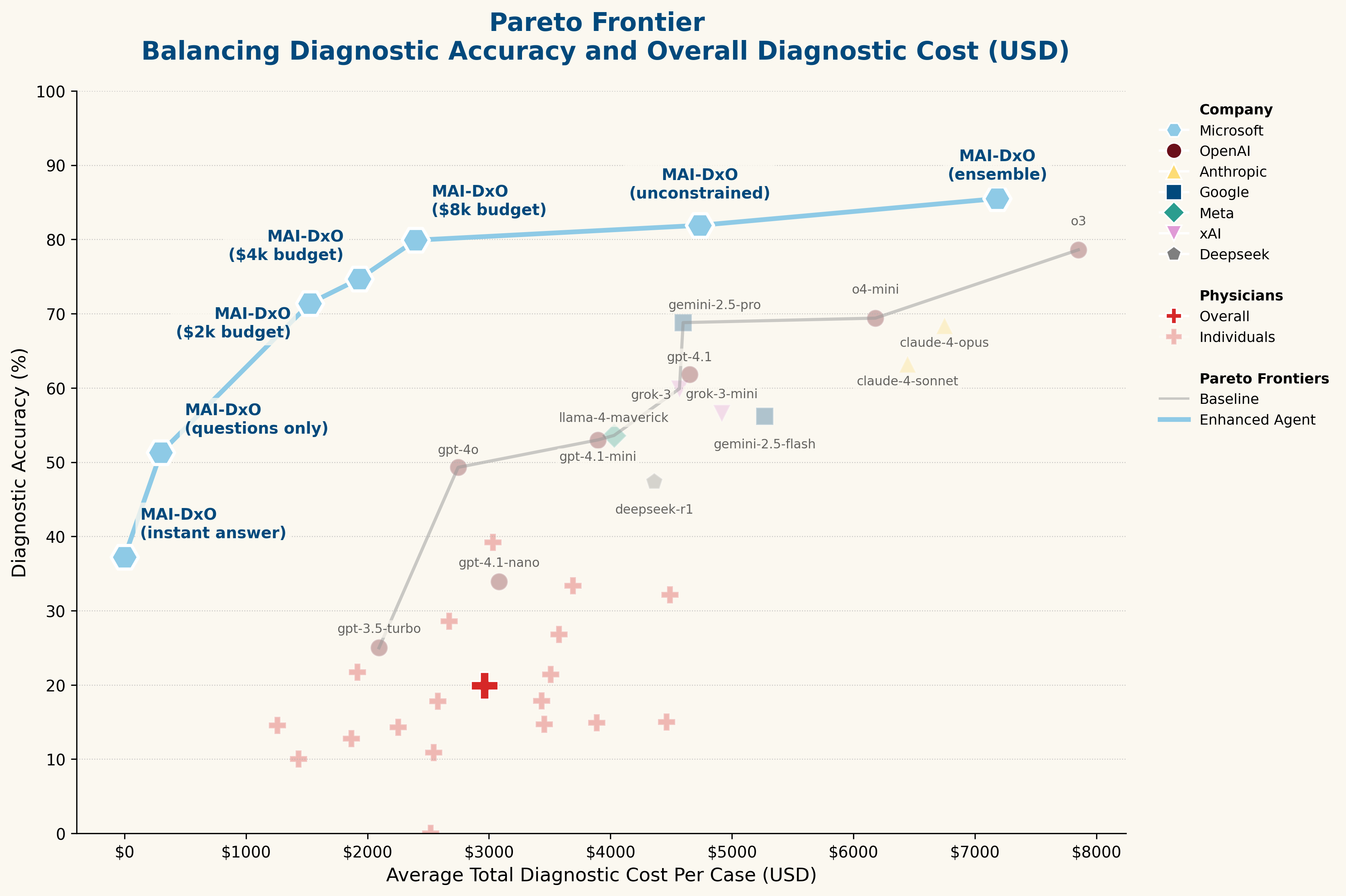}
    \caption{Pareto-frontier showing diagnostic accuracy versus average cumulative medical cost for each agent. Each test ordered on the diagnostic journey adds to total expenditure. Off-the-shelf models were evaluated using a uniform baseline prompt (see Figure \ref{fig:baseline-prompt}). \mai{}, built on top of the o3 model, achieves Pareto dominance over both off-the-shelf models and practicing physicians -- that is, at any price point, \mai{} achieves higher accuracy over other solutions.}
    \label{fig:pareto}
\end{figure}

\begin{figure}[H]
    \centering
    \includegraphics[width=\linewidth]{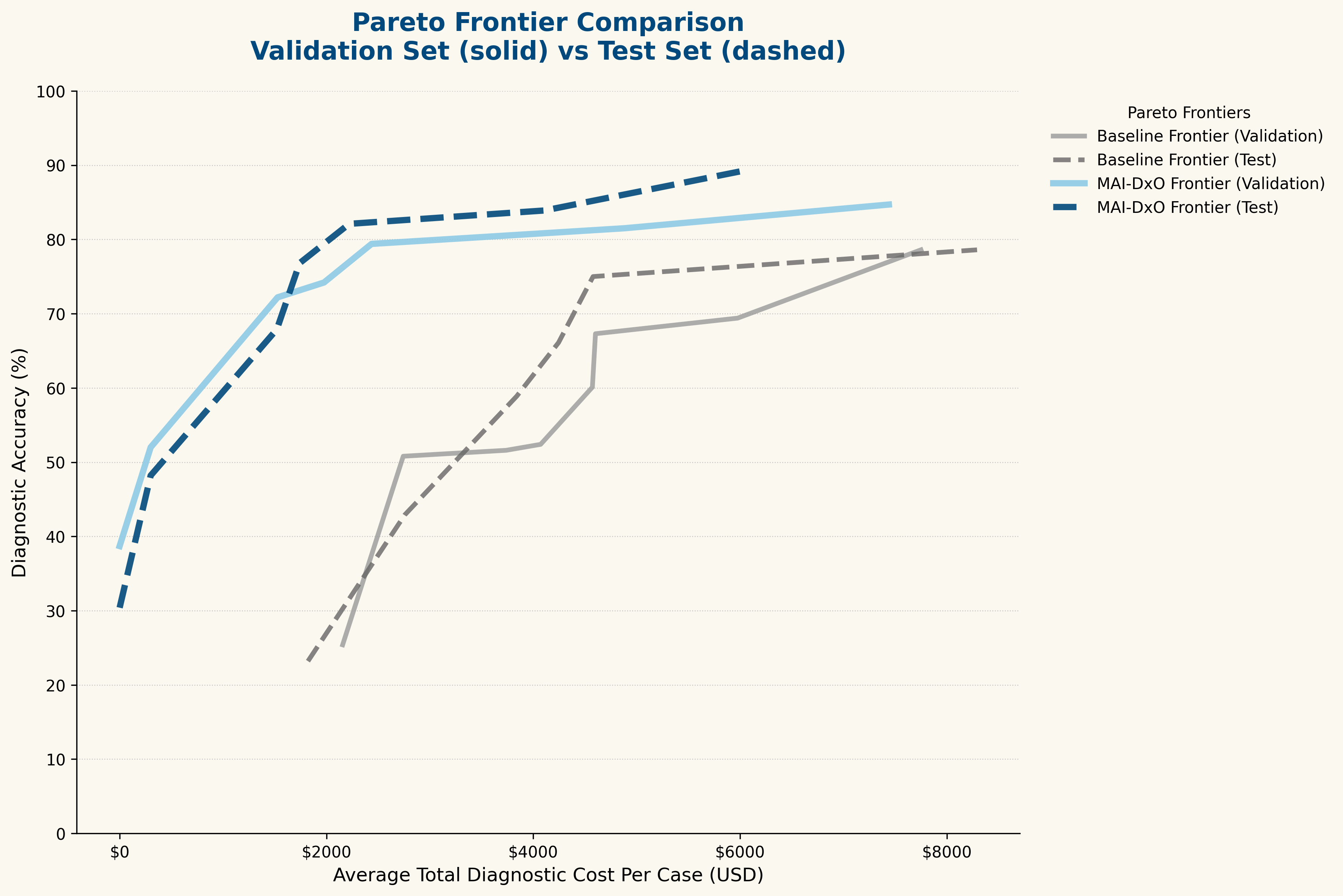}
    \caption{Pareto frontier curves of \mai{} and baseline prompting across validation and held-out test data. \mai{} continues to show significant improvement on performance on NEJM CPC cases published after model training cutoff, and thus provably outside the training corpora.}
    \label{fig:test-set-boost}
\end{figure}

We present the performance of all diagnostic agents on \nameShort{} in Figure \ref{fig:pareto}. Each point represents an AI agent's performance, with diagnostic accuracy plotted against average cumulative cost. AI agents are evaluated on all $304$ NEJM cases (including the $56$ test set cases), while physician performance is shown only for the held-out $56$ test set cases. Figure \ref{fig:test-set-boost} shows the corresponding Pareto frontiers computed on the test set, and indicates that AI agents tend to perform \emph{better} on this subset vs the $304$ cases.

\paragraph{Off-the-shelf model performance.} 

The Pareto frontier for off-the-shelf models ranged from modest accuracy (30-50\%) with minimal testing to 70-79\% accuracy with extensive testing (incurring \$4,000-7,900 in cost).
While some models dominated others (e.g. Gemini-2.5-Pro had higher accuracy than Claude-4 Sonnet and Opus, at \emph{less} cost), there was a correlation between diagnostic accuracy and cost, especially for reasoning models. Off-the-shelf o3 achieved the highest accuracy at 78.6\%, but also incurred the highest cost of \$7,850 per case.
This correlation indicates that information gathering remains crucial for diagnosis even for the most advanced AI systems, and supports the importance of information gathering as a key capability when evaluating models as diagnostic agents. Further, it suggests that more capable reasoning models are more skilled at assessing the complexity of cases, generating a broader set of differential diagnoses, and ordering more tests to confirm or exclude options.
Meanwhile, weaker models achieved a false economy by considering fewer possible differential diagnoses, thus never ordering the tests that would confirm or exclude them.

\paragraph{Physician performance.}

Each physician completed an average of 36 cases, providing a total of 764 human diagnoses over the 56 cases in the test set. On average, physicians spent 11.8 minutes per case, requesting 6.6 questions and 7.2 tests, resulting in an average cost of \$2,963. Their average diagnostic accuracy was 19.9\%. As with language models, we observed a correlation between diagnostic accuracy and cost incurred, although the variance for physicians is higher. Even the physician performing at the highest diagnostic accuracy (41\%) was outperformed by GPT-4o in both accuracy and cost. We note again that NEJM cases are selected for difficulty and instructional potential, so it is not surprising that physicians found them difficult.

\paragraph{MAI-Diagnostic Orchestrator performance.}

The blue curve in Figure \ref{fig:pareto} represents different operational modes of \mai{}, allowing for precise navigation of the accuracy-cost tradeoff. \mai{}, establishes a new Pareto frontier that outperforms off-the-shelf models \emph{and} physicians at every cost or accuracy level. We use o3 as the model for every \mai{} configuration throughout in Figure \ref{fig:pareto}, as model inference costs ($\approx$\$0.15 per case) are negligible compared to diagnostic test costs, making compute optimization largely irrelevant in the context of clinical diagnostic costs.

The standard \mai{} configuration (no budget) demonstrated particular effectiveness. When applied to o3, it achieved 81.9\% accuracy (vs off-the-shelf o3 at 78.6\%) while reducing average test costs to \$4,735 (from \$7,850).
This simultaneous improvement in both metrics illustrates that structured reasoning mitigates the accuracy-cost trade-off present in off-the-shelf models and physicians. As expected, adding budget constraints significantly lowered costs, with progressively more loss in accuracy as the budget was reduced. Notably, one configuration achieves 79.9\% accuracy at the low cost of \$2,396. On the other end of the spectrum, we were able to improve accuracy to $85.5\%$ with ensembling techniques, while still keeping the costs at \$7,184 (from o3's \$7,850).

We manually inspected a variety of cases where \mai{} (no budget) disagreed with off-the-shelf o3 with the baseline prompt. In these cases, \mai{} often sought disconfirming evidence and switched its diagnostic path (as a result of Dr. Hypothesis' explicit hypothesis tracking and Dr. Challenger's adversarial role), while off-the-shelf o3 seemed to anchor on initial impressions. Furthermore, the baseline seemed to lack a theory of information value, ordering tests that are ``reasonable'' given the current differential, rather than what maximally reduced diagnostic uncertainty per dollar spent. \mai{}'s Dr. Stewardship did not reject expensive tests outright, but forced the panel to ask whether the same information could be acquired at lower cost (in particular by asking the patient questions).
As an example, one particular case featured a patient hospitalized for alcohol withdrawal who ingested hand sanitizer, leading to intoxication. Off-the-shelf o3 fixated on antibiotic toxicity, ordering expensive imaging (including a brain MRI and EEG), and finally produced an incorrect diagnosis at a high cost of \$3,431. In contrast, Dr. Hypothesis flagged the need to consider
in-hospital toxin exposures given the timing in the very first round, and the panel asked about hand sanitizer ingestion before ordering tests. This direct question elicited the patient's confession, leading to targeted confirmatory testing (toxic alcohol panel showing elevated acetone) and a correct diagnosis at a total cost of only \$795.

\paragraph{\mai{} improved all off-the-shelf models.}
\begin{figure}[H]
    \centering    \includegraphics[width=\linewidth]{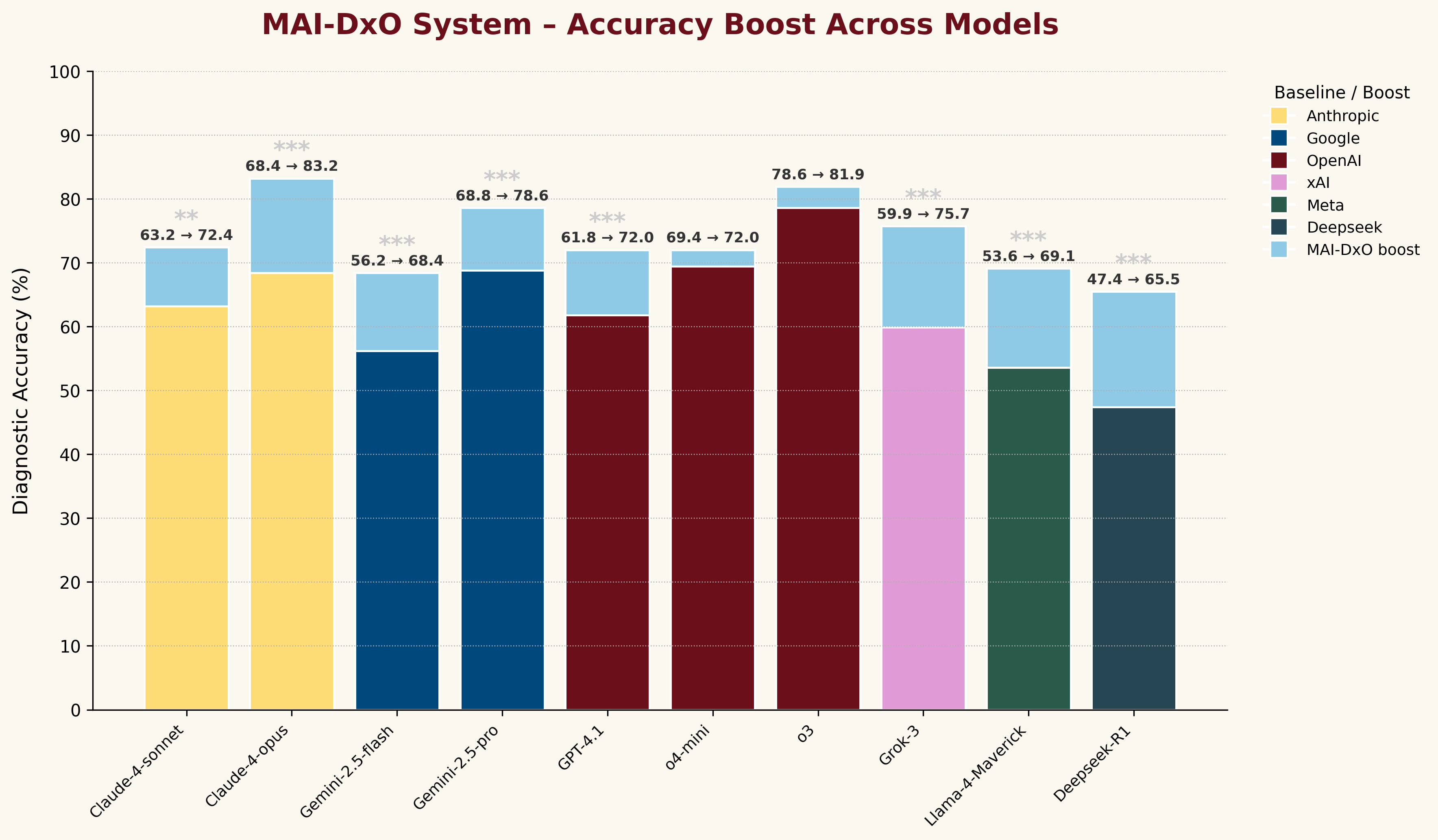}
    \caption{Accuracy improvements delivered by \mai{} (no budget constraints) across different large language models. Asterisks indicate statistical significance.}
    \label{fig:mai-boost}
\end{figure}

Even though \mai{} was primarily developed using GPT-4.1, its structured reasoning approach proved remarkably model-agnostic. Figure \ref{fig:mai-boost} demonstrates that \mai{} consistently improves diagnostic accuracy across all sufficiently capable foundation models, with particularly pronounced gains for weaker baselines, suggesting the framework helps weaker models overcome their limitations through structured reasoning. We computed the statistical significance of all accuracy gains in Figure \ref{fig:mai-boost} using a one-sided paired permutation test with $10000$ resamples. The gains for \mai{} (no budget) were statistically significant for all models ($p < 0.005$), except o3 and o4-mini which had very significant cost reductions over baseline ($p < 0.005$). Significant accuracy gains were also observed for o3 with \mai{} (ensemble) ($p < 0.005$). %

This convergence likely reflects how \mai{} compensates for different types of model weaknesses. For less capable models, the explicit maintenance of a differential diagnosis and systematic test selection provide scaffolding for medical reasoning they struggle with on their own. The virtual physician panel prevents common errors like premature closure or overlooking rare diseases. For more capable models, \mai{} appears to impose useful discipline—ensuring comprehensive differentials, reducing anchoring bias, and encouraging cost-conscious testing. Under simple, baseline prompting we hypothesize that models may rely on sets of inductive biases introduced during post-training for preparing them for general uses. Applying \mai{} may help override or reorient these inductive biases.

Interestingly, we found that raw diagnostic accuracy gains were more modest for OpenAI's reasoning models---likely because their baseline performance is already high. However, \mai{} was able to significantly improve their cost efficiency, as illustrated by the performance of o3 in Figure \ref{fig:pareto}. %

\paragraph{Results were robust across dataset splits.}

As noted earlier, the 56 most recent CPC cases, published between 2024 and 2025, were kept completely hidden from the development team as a ``test set", and no variants were run on them until methodologies were finalized. 
Partitioning data in this way is a common practice to measure and prevent overfitting, wherein a system fails to generalize beyond the data used during its training or validation. Strong performance on a truly held-out test set increases confidence on the system's ability to generalize.
While \mai{} does not update model weights (relying instead on prompting and orchestration), it is still possible for the system design choices to inadvertently overfit to cases employed during its validation. This particular train-test split also checks against potential memorization. While the NEJM cases are hidden behind a paywall, it is still possible that some off-the-shelf models were trained on them in some way. However, the majority of cases in the test set occurred \emph{after the training data cutoff} of the models we report on. 

In Figure \ref{fig:test-set-boost}, we report stratified Pareto frontier curves of model performance across the validation (248 cases) and test (56 cases) sets. The \mai{} system exhibited comparable absolute performance on the test set, with the relative improvements over off-the-shelf models preserved in both diagnostic accuracy and cost efficiency. These results suggest that the performance gains are robust and not driven by memorization effects.

\section{Discussion}

We introduce \nameShort{}, a benchmark that transforms 304 New England Journal of Medicine CPC cases into interactive, multi-turn diagnostic challenges. Unlike static medical benchmarks that present all information upfront, \nameShort{} more closely mirrors real-world clinical practice: diagnosticians start with minimal information and must actively decide which questions to ask, which tests to order, and when to issue a final diagnosis, with each decision incurring realistic costs. Through careful engineering, including a Gatekeeper that can synthesize plausible results for tests not described in the original cases and a clinically validated Judge to assess diagnostic accuracy, we introduce a robust evaluation environment for sequential clinical reasoning.

Within this framework, we present \mai{}, a system that simulates panels of different clinical personas in order to decide which questions or tests to request. \mai{} significantly improved diagnostic accuracy beyond strong off-the-shelf models, while simultaneously reducing cumulative test costs in \nameShort{}, thereby establishing a new Pareto frontier between accuracy and medical cost.

\subsection{Explaining superhuman performance}

When doctors begin their careers, they face a key decision: should they become generalists, with broad knowledge across many medical areas, or specialists, with deep expertise in a narrow field? This division is necessary because medicine is too vast for any one person to master in full. To manage this complexity, healthcare systems rely on collaboration: generalists and specialists work together in clinics and hospitals, combining their diverse and complementary knowledge and decision-making skills to provide patients with the comprehensive and effective care that they need.

Today, frontier AI language models are challenging this traditional structure. These advanced systems show remarkable versatility, demonstrating both broad and deep medical understanding, and the polymathic ability to reason across specialties. In effect, they combine the generalist’s range with specialists' depth. As a result, they significantly outperform individual physicians on complex diagnostic problems, such as those featured in the NEJM CPC cases. Our findings highlight this impressive capability. Expecting any single doctor to master the full range of such cases is unrealistic.   

Consider, for example, a complex undiagnosed cancer case. A primary care physician's role is to generate initial hypotheses and to refer the patient to the appropriate oncology specialist for further diagnostic workup. The specialist then oversees advanced diagnostic tests to reach a conclusive diagnosis---steps that the generalist would not typically manage.

This raises an intriguing question: When evaluating frontier AI systems, should we evaluate frontier AI systems by comparing them to individual physicians, or to entire hospital-like teams of generalists and specialists? The answer to this question will help both define and shape the future role of AI in healthcare.

\subsection{Related Work}
Medical problem solving has been a longstanding field of study within the medical community. In the medical AI literature, sequential diagnosis was formalized several decades ago through \emph{normative} models grounded in Bayesian probability and decision theory \citep{decisiontheory88}. This framework enabled expert-level sequential diagnostic systems in domains such as nephrology \citep{gorry1968experience}, pathology \citep{horvitzetal1984,heckerman1992toward}, and trauma care \citep{horvitz1997time}. However, widespread adoption was hindered by the practical challenges of engineering these systems, particularly bottlenecks around the need to acquire detailed, expert-curated data on probabilistic relationships between findings and diseases. 

More recent work has shifted toward the application of LMs to medical challenge problems, which typically include clinical reasoning as part of a broader evaluation suite \citep{biomedbert, pubmedbert, chatgptusmle, gpt4usmle, multimedqa, bedi2025medhelm}. While these studies demonstrated foundational performance leaps at their time of publication, existing multiple-choice benchmarks have now become saturated, highlighting the need for more complex and realistic assessments, as well as careful end-to-end agent optimization in healthcare tasks \citep{bedi2025optimization}.

To this end, there have been multiple studies, notably the \emph{Articulate Medical Intelligence Explorer} (AMIE) line of work, which leveraged NEJM content as source material for challenging benchmarks. For diagnostic capability assessments, AMIE also leveraged NEJM-CPC cases; however, this line of work assessed models in a fixed "vignette" style setting in which the case information was summarized into a compact prompt and the models were asked to make a top-10 differential diagnosis \citep{mcduff2025towards}. In contrast, our key differentiation was to transform the static clinical case information into the real-world evidential reasoning challenge characterized by sequential diagnosis, which assesses models on their ability to iteratively ask for information, \textit{starting from minimal information}, in a cost-sensitive manner and decide when a diagnosis should be made. Of note, in a parallel paper \citep{tu2025towards} AMIE was also assessed on conversational quality dimensions, such as empathy. While these represent critical dimensions of interaction with physicians and patients, we chose to frame physicians' and agents' interaction with SDBench as an interaction with an ``oracle'' about the patient, and so primarily focused on measures of cost and diagnostic accuracy. We note that \citep{li2024mediq} also tests language models on information gathering capabilities; however, this work builds on much simpler, multiple-choice USMLE-style questions (which are a few sentences long; by contrast, NEJM CPC cases are several pages long). The authors also focus purely on information gathering via patient questions; in this work, we enable the extra dimension of ordering diagnostic tests and measuring cost efficiency. 

More recently, \citep{brodeur2024superhuman} utilized cases from NEJM's Healer Platform, a digital platform designed to teach and assess clinical reasoning through interactive, case-based learning, in addition to NEJM-CPC cases. Notably, the cases within the NEJM Healer platform are designed to serve as educational aids for healthcare professionals in training and do not pose the same diagnostic challenge as the NEJM-CPC cases. As with \cite{mcduff2025towards}, the presenting information from the NEJM-CPC cases were presented as fixed vignettes to a model, which generated a differential diagnosis and the next most appropriate diagnostic test. Similarly, \citep{schmidgall2024agentclinic} leverages the NEJM Image Challenges, which are multiple choice image question and answer tasks. 

\subsection{Limitations}

Since \nameShort{} is built from complex, pedagogically curated NEJM CPC cases, the case distribution does not match that of a real-world deployment scenario, and indeed there are no cases where the patients are in fact healthy or have benign syndromes. Thus, we do not know whether \mai{}'s performance gains on hard cases generalize to common, everyday clinical conditions, and could not measure false positive rates. Additionally, a practical diagnostic agent must incorporate patient-specific risk factors, and consider additional factors beyond cost, e.g. invasiveness and risk to patients, patient discomfort and wait times, expected delays before receiving results in the face of acute illness, availability of tests at current location, and constraints of authorization and reimbursement.

While our estimates of medical costs reflected test costs in the United States, in reality costs vary across geography, health systems, payers, and providers. Further, there are costs beyond the tests themselves, such as physician reporting time, device maintenance, patient travel costs for tests, etc. While recognizing that our medical cost estimates are best viewed as first-order approximations, they are consistent across all agents, and thus still help quantify relative trade-offs between accuracy and resource use.

While our report of physician performance is useful in comparing humans to AI diagnostic systems, it is also meant as a first-order approximation.
Given the breadth of diagnoses represented within NEJM CPCs, we opted to recruit medical generalists only (primary care physicians and internal physicians), while in reality these might refer more complex cases to specialists. Further, we asked the participants in our study to refrain from using search engines (to prevent them finding the exact NEJM cases online), while in reality physicians are free to use such tools, including electronic medical records that often contain care guidelines, consult colleagues or textbooks, or even off-the-shelf LMs. While acknowledging these limitations, our results indicate possible accuracy gains, especially when considering clinicians working in remote and under-resourced settings, and also give us a picture of how LMs could augment medical expertise to improve health outcomes even in well-resourced settings.

\subsection{Implications and Future Work}

Our findings demonstrate the promise of AI methods for sequential diagnosis, including the ability to explicitly model working differential diagnoses and reason about informational value and cost of diagnostic tests. While these results do not yet establish the clinical efficacy of \mai{} in real-world decision support, they underscore AI's increasing potential to address urgent challenges in healthcare delivery.
Our model-agnostic system design may alleviate risks and implementation challenges for health systems aiming to adopt best-in-class language model–based diagnostic support in a rapidly evolving field. By reducing reliance on any single model, it avoids the need to ``version chase'' each new model release.
In terms of practical application, future work should validate \mai{} in everyday clinical environments, where disease prevalence and presentations reflect routine practice rather than the rare, complex cases featured in the NEJM CPC corpus. An immediate goal is to identify the settings in which \mai{} could address unmet needs and deliver the greatest value to health outcomes and societal benefit. 

We hypothesize that access to superhuman diagnostic capabilities requiring minimal health IT infrastructure could improve quality of care globally, helping to mitigate the costly impact of clinical workforce shortages and variability in care delivery \cite{wennbergimproving, mandl2025ai}. In resource-limited settings especially, cost-effective strategies may enable health systems to impact more lives per dollar spent, allowing scarce medical resources to be reserved for those with the most urgent clinical needs. More broadly, such systems might even make direct-to-consumer tools possible, such as smartphone-based triage, provided that safety, regulatory clearance, and data-privacy safeguards are demonstrably in place.

Progress toward effective clinical decision support will require the development of diagnostic corpora that mirror real-world prevalence patterns. Such benchmarks will help to surface limitations and opportunities for refinement that may be obscured by our current emphasis on especially difficult diagnostic scenarios.  Second, our synthetic findings framework could support the development of large-scale interactive medical benchmarks beyond the 304 cases available here. Beyond evaluation of AI systems, the methodology we have developed could be used to enhance medical education and training, enabling students and practitioners to practice diagnostic reasoning in simulated interactive environments, potentially guided by AI-based pedagogical support. Finally, incorporating visual and other sensory modalities, such as imaging, could push diagnostic accuracy even higher while maintaining cost efficiency.

\section*{Code and data availability}

We are in the process of submitting this work for external peer review and are actively working with partners to explore the potential to release \nameShort{} as a public benchmark. 

\section*{Acknowledgments}

We are grateful to NEJM Group for permission to use the NEJM cases in the research reported in this paper.
The research described here has benefited from work across many teams at Microsoft.  We thank esteemed colleagues both inside and outside of Microsoft for sharing their insights including Bryan Bunning, Nando de Freitas, Kenneth D Mandl, Andrija Milicevic, Joseph Petro, Hoifung Poon, David Rhew, Adam Rodman, Nigam Shah, Ted Shortliffe, Karén Simonyan, Eric Topol, Bob Wachter, and Jim Weinstein.
Gianluca Fontana and Kevin Hawkins (Prova Health) provided support on the health economics and outcomes section.

\bibliographystyle{plainnat}
\bibliography{mainbib}

\newpage

\appendix
\section{Batched vs Single Testing}

The main results for \mai{} and baseline agents were all performed using batch testing mode, where the agents were allowed to ask many questions or request many tests within one turn. \mai{} agents were instructed to limit this to 5 questions and 3 tests. Clinician review of our benchmark setup suggested that batching questions and tests reflected  more realistic testing strategies, to alleviate patient discomfort, reduce test result wait times and for practical reasons concerning laboratory test processing. However, for usability reasons the human clinicians used the UI-version of the benchmark in a single-turn fashion. We ran an ablation on the test set only for \mai{} (o3) to compare single vs batched mode. We found that both single and batched mode agents had identical accuracy (83.9\%), but the single-mode agent actually had a lower cost (\$3,991 vs \$5,084). However, to preserve the aforementioned realism, we report \mai{} results in the batch setting.

\section{Visualizing Model and Clinician Performance}

\begin{figure}[H]
    \centering
    \includegraphics[width=1\linewidth]{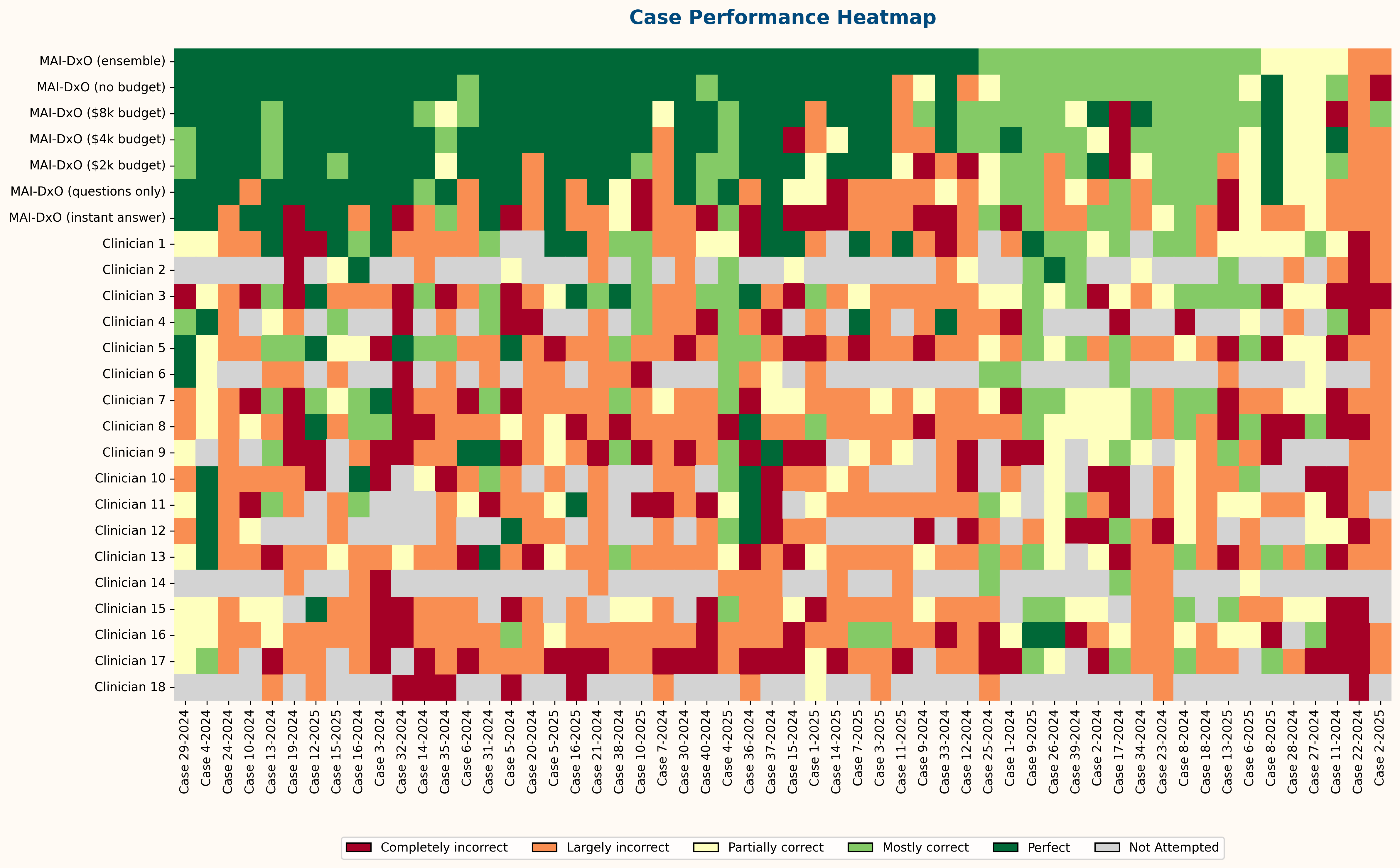}
    \caption{Case level scores of \mai{} variants and clinicians across the 56 case test set. All clinicians who assessed at least 10 cases are shown. Model variants and clinicians are ordered by their diagnostic accuracy and cases are ordered by their perceived difficulty according to  \mai{} (ensemble).}
    \label{fig:model-clinician-heatmap}
\end{figure}

\section{Recent CPC Case}

The most recently published CPC case, ``Case 18-2025: A 63-Year-Old Woman with Dyspnea on Exertion", was published on June 26, 2025.

\mai{} was run on the case on June 26, 2025 and the transcript between \mai{} and the Gatekeeper for the case is shown on the pages below. Note that the transcript shows indicators of which test results were produced from the case text, and which ones were generated synthetically for illustrative reasons -- this information is not revealed in the interface used by models and physicians attempting to solve the case. 



\section*{Supplementary material}

# Case 504: "Case 18-2025: A 63-Year-Old Woman with Dyspnea on Exertion"

## Agent Dialog (14 messages)

**GATEKEEPER**

A 63-year-old woman was evaluated for dyspnea and edema. Soft-tissue thickening was noted in the retroperitoneum and mediastinum and around the heart and vessels. A diagnosis was made.

**MAI-DXO**

<question>Can you describe in your own words the main problems that brought you in today and when each one began?</question>
<question>How have your breathlessness and body swelling changed over time—what was the exact onset, how rapidly did they progress, what makes them better or worse, and how severe are they now?</question>
<question>Besides shortness of breath and edema, have you noticed any chest pain, cough, palpitations, fever, night sweats, weight change, abdominal discomfort, change in urine or bowel habits, or any other new or unusual symptoms?</question>
<question>Please list every prescription medication, over-the-counter drug, vitamin, supplement, or herbal product you take, including dosages and the approximate start dates for each. </question>
<question>Have you undergone any surgeries, hospitalizations, implanted device placements, catheterizations, or other medical procedures—recently or in the past—that we should know about? </question>

**GATEKEEPER**

Test Results:
1. Can you describe in your own words the main problems that brought you in today and when each one began? REAL : Primary complaints: progressive dyspnea on exertion for 2 months, dry cough for approximately 6 weeks, orthopnea requiring three pillows, lower-extremity edema, abdominal distention, anorexia, postprandial

nausea, and a 2-month history of dry mouth, polydipsia, and polyuria.

2. How have your breathlessness and body swelling changed over time—what was the exact onset, how rapidly did they progress, what makes them better or worse, and how severe are they now? REAL : Dyspnea began about 2 months ago during exertion, progressed over 2 weeks to occur with minimal activity, and now limits routine exercise and sleep. Orthopnea developed concurrently. Lower-extremity and abdominal swelling appeared around the same time, worsened over weeks, and partially improved with increased diuretic dosing; current edema is significant in the lower legs.

3. Besides shortness of breath and edema, have you noticed any chest pain, cough, palpitations, fever, night sweats, weight change, abdominal discomfort, change in urine or bowel habits, or any other new or unusual symptoms? REAL : Reports dry cough worse when supine and intermittent palpitations described as "fast heart rates." Denies chest pain or pressure, fever, chills, night sweats, weight change, abdominal pain, back or flank pain, hematuria, dysuria, rash, photosensitivity, hair or nail changes, or neurologic symptoms.

4. Please list every prescription medication, over-the-counter drug, vitamin, supplement, or herbal product you take, including dosages and the approximate start dates for each. SYNTHETIC : Metformin 500 mg orally twice daily (started ~5 years ago); calcium carbonate 600 mg orally once daily (started ~2 years ago); daily multivitamin (started ~5 years ago); furosemide 40 mg orally once daily (started ~7 weeks ago); colchicine 0.6 mg orally twice daily (started ~1 year ago); desmopressin 0.2 mg orally at bedtime (started ~8 days ago).

5. Have you undergone any surgeries, hospitalizations, implanted device placements, catheterizations, or other medical procedures—recently or in the past—that we should know about? SYNTHETIC : Surgical resection of a meningioma (several years ago); knee arthroscopy (approximately 10 years ago); glucocorticoid injections for shoulder adhesive capsulitis (6 months ago); closed reduction and casting for right distal radius fracture (4 weeks ago). No implanted devices or indwelling catheters reported.

MAI-DXO

<question>Have you experienced dryness or grittiness in your eyes, enlarged or painful salivary glands, difficulty swallowing dry foods, or recurrent dental cavities?</question>

<question>Do you have joint pains, morning stiffness, muscle weakness, skin rashes (especially on the face or hands), Raynaud's phenomenon, photosensitivity, or episodes of swelling or pain in the parotid glands?</question>

<question>Please describe your occupation, hobbies, recent travel, home environment (molds, new construction), and any exposures to chemicals, dusts, animals, or infectious contacts.</question>

<question>Do you smoke cigarettes or use tobacco, drink alcohol, or

use recreational drugs? If yes, how much and for how long?
</question>
<question>Does anyone in your family have autoimmune diseases, heart or kidney disease, cancer, or similar symptoms of swelling and shortness of breath?</question>

**GATEKEEPER**

Test Results:

1. Have you experienced dryness or grittiness in your eyes, enlarged or painful salivary glands, difficulty swallowing dry foods, or recurrent dental cavities? `SYNTHETIC` : Patient reports a 2-month history of dry mouth but denies dryness or grittiness in the eyes, salivary gland enlargement or pain, difficulty swallowing dry foods, or recurrent dental cavities.

2. Do you have joint pains, morning stiffness, muscle weakness, skin rashes (especially on the face or hands), Raynaud's phenomenon, photosensitivity, or episodes of swelling or pain in the parotid glands? `SYNTHETIC` : Patient reports ongoing hand-joint stiffness with swan-neck deformities similar to her baseline rheumatoid arthritis. She denies morning stiffness beyond her usual baseline, muscle weakness, skin rashes, Raynaud's phenomenon, photosensitivity, or swelling or pain in the parotid glands.

3. Please describe your occupation, hobbies, recent travel, home environment (molds, new construction), and any exposures to chemicals, dusts, animals, or infectious contacts. `SYNTHETIC` :
Patient works in an office, lives with her husband in a wooded area of New England, and has no recent travel. She reports no known exposures to molds, new construction, chemicals, dusts, animals, or infectious contacts.

4. Do you smoke cigarettes or use tobacco, drink alcohol, or use recreational drugs? If yes, how much and for how long? `REAL` :
Patient denies tobacco or recreational drug use and reports drinking three glasses of wine per week.

5. Does anyone in your family have autoimmune diseases, heart or kidney disease, cancer, or similar symptoms of swelling and shortness of breath? `REAL` : Family history is notable for coronary artery disease and colon cancer in her father, systemic lupus erythematosus in her son, and hypertension and diabetes mellitus in multiple other relatives.

**MAI-DXO**

<question>Can you estimate your daily urine volume (for example, number of times you void and approximate volume each time) and explain why desmopressin was prescribed 8 days ago—was a diagnosis such as diabetes insipidus made?</question>
<question>Have you had any imaging studies (CT, MRI, ultrasound) of the chest, abdomen, or pelvis in the past few months? If so, what

findings were reported, especially regarding the pancreas, kidneys, aorta, or lymph nodes?</question>
<question>Have you noticed any abdominal or back pain, flank discomfort, early satiety, jaundice, change in stool color, itchy skin, or episodes of pancreatitis?</question>
<question>Have you ever been told you had elevated liver enzymes, abnormal kidney function, or protein in your urine? If yes, what were the results and when?</question>
<question>Do you currently have or have you ever had painless swellings or lumps around your eyes, salivary glands (parotid/submandibular), thyroid, or any other soft-tissue mass noticed on exam or imaging?</question>

**GATEKEEPER**

Test Results:
1. Can you estimate your daily urine volume (for example, number of times you void and approximate volume each time) and explain why desmopressin was prescribed 8 days ago—was a diagnosis such as diabetes insipidus made? `SYNTHETIC` : Patient reported approximately 8–10 voids per day, with volumes of about 300–400 ml per void, resulting in an estimated total daily urine output of 2.5–3 L. Desmopressin was prescribed 8 days ago after endocrinology confirmed central diabetes insipidus.
2. Have you had any imaging studies (CT, MRI, ultrasound) of the chest, abdomen, or pelvis in the past few months? If so, what findings were reported, especially regarding the pancreas, kidneys, aorta, or lymph nodes? `REAL` : Underwent CT of the abdomen and pelvis 8 days ago showing anterior pararenal and bilateral perirenal fluid and soft-tissue thickening; small pericardial effusion; hepatic cysts; uterine leiomyoma; bilateral pleural effusions. One week later, abdominal MRI showed extensive hypointense soft-tissue thickening in the retroperitoneal space surrounding both kidneys, the abdominal and descending thoracic aorta, and a bulky pancreatic head; pleural effusions and pericardial thickening were also noted. CT of the chest with contrast demonstrated pulmonary edema; bilateral pleural effusions; small pericardial effusion with pericardial thickening; soft-tissue thickening surrounding the ascending and descending thoracic aorta.
3. Have you noticed any abdominal or back pain, flank discomfort, early satiety, jaundice, change in stool color, itchy skin, or episodes of pancreatitis? `SYNTHETIC` : Denies abdominal pain, back or flank discomfort, early satiety, jaundice, changes in stool color, pruritus, or any episodes of pancreatitis.
4. Have you ever been told you had elevated liver enzymes, abnormal kidney function, or protein in your urine? If yes, what were the results and when? `SYNTHETIC` : Denies any prior notification of elevated liver enzymes, abnormal kidney function, or proteinuria.
5. Do you currently have or have you ever had painless swellings or

lumps around your eyes, salivary glands (parotid/submandibular), thyroid, or any other soft-tissue mass noticed on exam or imaging?
**SYNTHETIC** : Denies any painless swellings or lumps around the eyes, salivary glands, thyroid, or other soft-tissue masses noted on exam or imaging.

**MAI-DXO**

Lead-Dx Summary
1 IgG4-related disease　2 Erdheim-Chester disease　3 Infiltrative lymphoma.
Decisive test: Serum IgG4—marked elevation strongly supports IgG4-RD and is inexpensive.

Dr. Test-Chooser: Add CBC & metabolic panel to screen for cytopaenia, renal/liver dysfunction (safety-net) and ESR/CRP for inflammatory burden; all cheap, single venipuncture.

Dr. Challenger: Bone involvement could flip to Erdheim-Chester; but serum IgG4 first is quicker & cheaper.

Dr. Stewardship: No duplicates, ≤3 tests, all low-cost—approved.

<test>Serum IgG4 concentration</test>
<test>Complete blood count with differential & basic metabolic panel</test>
<test>ESR and CRP</test>

**GATEKEEPER**

Test Results:
1. Serum IgG4 concentration　**SYNTHETIC** : Serum IgG4 concentration: 62 mg per deciliter (reference range 4–86 mg per deciliter)
2. Complete blood count with differential & basic metabolic panel
**REAL** : White-cell count: 7150 per µl (normal); differential count: normal; sodium: 139 mmol/L; potassium: 4.9 mmol/L; chloride: 105 mmol/L; carbon dioxide: 23 mmol/L; urea nitrogen: 13 mg/dL; creatinine: 0.50 mg/dL; glucose: 104 mg/dL
3. ESR and CRP　**REAL** : Erythrocyte sedimentation rate: 70 mm/hr; C-reactive protein: 47.1 mg/L

**MAI-DXO**

Lead-Dx Summary
1 Erdheim-Chester disease　2 Infiltrative lymphoma　3 IgG4-related disease.
Decisive test: Perirenal soft-tissue biopsy for histology/IHC—foamy CD68+ CD1a– histiocytes would prove Erdheim-Chester.

Dr. Test-Chooser: Add cheap bilateral long-bone (femur–tibia) X-rays—typical symmetrical diaphyseal osteosclerosis is highly suggestive. Reflex BRAF V600E on the same biopsy confirms molecularly and guides therapy.

Dr. Challenger: If biopsy shows lymphoma features, lead flips; bone films still discriminate as lymphoma seldom causes classic ECD bone pattern.

Dr. Stewardship: All unique, ≤3 tests, biopsy essential per tissue-proof rule—approved.

<test>CT-guided perirenal (or periaortic) soft-tissue core biopsy with histopathology and immunohistochemistry (CD68, CD1a, S100, IgG4, factor XIIIa)</test>
<test>Plain radiographs of bilateral femurs and tibias (antero-posterior and lateral)</test>
<test>BRAF V600E mutation analysis on biopsy tissue (PCR or IHC)</test>

**GATEKEEPER**

Test Results:
1. CT-guided perirenal (or periaortic) soft-tissue core biopsy with histopathology and immunohistochemistry (CD68, CD1a, S100, IgG4, factor XIIIa) SYNTHETIC : Histologic sections of core biopsy show a proliferation of histiocytoid cells with small ovoid nuclei and abundant eosinophilic to foamy cytoplasm in a fibrous stroma. Thin-walled vessels and scattered lymphocytes are present; no granulomas or giant cells are seen. Immunohistochemical staining: CD68 positive; CD1a negative; S100 negative; IgG4 shows fewer than 5 IgG4-positive plasma cells per high-power field with an IgG4/IgG ratio of <10%; factor XIIIa positive in lesional cells.
2. Plain radiographs of bilateral femurs and tibias (antero-posterior and lateral) SYNTHETIC : Plain radiographs of bilateral femurs and tibias (AP and lateral) show symmetric diaphyseal cortical and medullary sclerosis without lytic lesions or periosteal new bone formation. Joint spaces are preserved.
3. BRAF V600E mutation analysis on biopsy tissue (PCR or IHC) REAL : Immunohistochemical staining for BRAF V600E shows diffuse positivity in lesional histiocytes. Rapid molecular assay on formalin-fixed paraffin-embedded tissue confirms the presence of a BRAF V600E/D mutation.

**MAI-DXO**

Erdheim-Chester disease

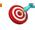

**MAI-DXO DIAGNOSIS CONFIRMATION**

Are you quite sure about your diagnosis?

Consider:
- Are there any additional tests that could confirm or rule out your diagnosis?
- Have you considered all relevant differential diagnoses?
- Are there any specialized tests that would be helpful?
- Any follow-up questions that could provide more clarity?

Prefer questions or cheaper / less invasive tests if possible, but feel free to order more expensive tests if you think they are necessary to _confirm_ or _rule out_ your diagnosis.
Don't ask for any test you have already ordered though. Before asking for a test, ask yourself: 'Have I already ordered this test or another test that would be redundant?'

If you are not sure, continue using <question></question> tags for questions and <test></test> tags for tests.
If you are sure, just provide your diagnosis using <diagnosis></diagnosis> tags again.

**MAI-DXO**

<diagnosis>Erdheim-Chester disease</diagnosis>

**Model Diagnosis:**

Erdheim-Chester disease

**Actual Diagnosis:**

Erdheim-Chester disease.



\end{document}